\documentclass[10pt,conference]{IEEEtran}
\IEEEoverridecommandlockouts
\usepackage{cite}
\usepackage{amsmath,amssymb,amsfonts}
\usepackage{algorithmic}
\usepackage{graphicx}
\usepackage{textcomp}
\usepackage{subfig}
\usepackage{xcolor}
\usepackage{enumitem}
\usepackage{float}
\usepackage{multirow}

\DeclareMathOperator*{\argmax}{arg\,max}
\def\BibTeX{{\rm B\kern-.05em{\sc i\kern-.025em b}\kern-.08em
    T\kern-.1667em\lower.7ex\hbox{E}\kern-.125emX}}
\begin{document}

\title{Enhancing Efficiency of Quadrupedal Locomotion over Challenging Terrains with Extensible Feet\\
{\footnotesize \textsuperscript{}
\thanks{* Lokesh Kumar and Sarvesh Sortee contributed equally to this work.}}
}

\author{\IEEEauthorblockN{Lokesh Kumar$^*$}
\IEEEauthorblockA{
\textit{TCS Research}\\
Kolkata, India \\
kumar.lokesh7@tcs.com}
\and
\IEEEauthorblockN{Sarvesh Sortee$^*$}
\IEEEauthorblockA{
\textit{TCS Research}\\
Kolkata, India \\
sarvesh.sortee@tcs.com}
\and
\IEEEauthorblockN{Titas Bera}
\IEEEauthorblockA{
\textit{TCS Research}\\
Kolkata, India \\
titas.bera@tcs.com}
\and
\IEEEauthorblockN{Ranjan Dasgupta}
\IEEEauthorblockA{
\textit{TCS Research}\\
Kolkata, India \\
ranjan.dasgupta@tcs.com}
}

\maketitle

\begin{abstract}

Recent advancements in legged locomotion research have made legged robots a preferred choice for navigating challenging terrains when compared to their wheeled counterparts. This paper presents a novel locomotion policy, trained using Deep Reinforcement Learning, for a quadrupedal robot equipped with an additional prismatic joint between the knee and foot of each leg. The training is performed in NVIDIA Isaac Gym simulation environment. Our study investigates the impact of these joints on maintaining the quadruped's desired height and following commanded velocities while traversing challenging terrains. We provide comparison results, based on a Cost of Transport (CoT) metric, between quadrupeds with and without prismatic joints. The learned policy is evaluated on a set of challenging terrains using the CoT metric in simulation. Our results demonstrate that the added degrees of actuation offer the locomotion policy more flexibility to use the extra joints to traverse terrains that would be deemed infeasible or prohibitively expensive for the conventional quadrupedal design, resulting in significantly improved efficiency.
\end{abstract}


\section{Introduction}
Legged locomotion in environments with non-flat unstructured terrain is a complex task. Traditional control methods used for such problems tend to be composed of several different interconnected modules, e.g. a perception processing module, a footstep planning module, a state estimator module, a gait/trajectory generator module and reactive modules for reflex control in case of foot slippage \cite{griffin2019footstep,bledt2018contact,focchi2018slip,camurri2017probabilistic}. These architectures become increasingly complex as more difficult operating conditions are considered. Other approaches such as Deep Reinforcement Learning (DRL), especially in model free RL, have seen rapid development as an alternative to these traditional methods. In DRL the robots learn to take decisions using a control policy based on the rewards obtained from continuous interactions with the simulation environments. In this paper we present a novel approach of locomotion in challenging terrains by making the robots feet extensible. This is done by adding prismatic joints after the knee. The motivation behind adding these joints is to add capability for traversing the terrains which have height and depth greater than the reach of a conventional quadruped leg, and also to use these extra degrees of freedom to maintain a desired height. This can be helpful for tasks such as creating a terrain map inside a coal mine or tunnel, or scenarios where the surface is not firm such as mud and snow, where the quadruped have to maintain a specific body height. To the best of our knowledge no similar study has been done in the present literature, where the effect of such modification are evaluated thus making the proposed work novel. The details of the work is discussed in upcoming sections.

\begin{figure}[t]
\centerline{\includegraphics[scale=0.375]{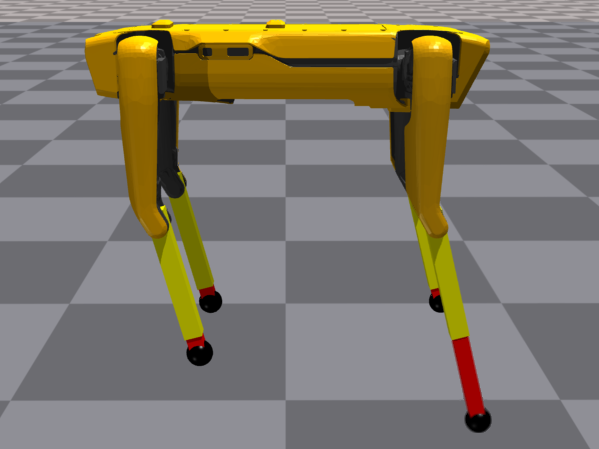}}
\caption{Quadrupedal robot with an actuated prismatic joint between the knees and feet, depicted here by the extension of the \textit{red} link}
\label{robot_image}
\end{figure}

\subsection{Related work}

Walking on irregular terrain poses a complex challenge for legged locomotion, and various methods have been developed to address this problem. Early works such as \cite{kalakrishnan2010fast,kolter2008control} considered terrain traversal as a solution to a  hierarchical planning problem, choosing a foothold followed by planning a leg trajectory to that foothold, between every foot-terrain contact event. Central Pattern Generators (CPG) have been used in combination with sensory data to enable quadrupeds to traverse unforeseen conditions \cite{fukuoka2003adaptive}. Righetti et al. \cite{righetti2008pattern} used feedback from touch sensors placed on the feet to change leg motions and traverse given terrains, while Hyun et al. \cite{hyun2014high} used a hierarchical controller combined with proprioceptive impedance control. S. Gay et al. \cite{gay2013learning} used feedback from the camera and the gyroscope combined with a CPG to learn walking over different terrains. Model Predictive Control (MPC) approaches use simplified dynamic models and solve an online optimization problem at every timestep to achieve a desired motion. \cite{grandia2022perceptive} presents a perception coupled MPC approach for dynamic locomotion over unstructured terrain. One disadvantages of these approaches is that a gait specification has to be given externally and reactive behaviours might need to be predefined to tackle slippage and incorrect contact estimation.

\begin{figure*}[tbh]
\centering
\subfloat[\textit{SmoothSlope}]{\label{4figs-a} \includegraphics[width=0.3\linewidth, height=2.5cm]{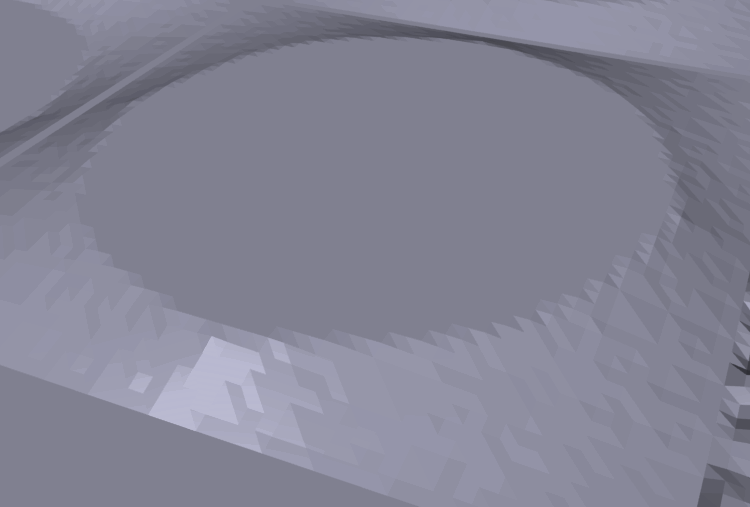}}
\hfill
\subfloat[\textit{RoughSlope}]{\label{4figs-b} \includegraphics[width=0.3\linewidth,height=2.5cm]{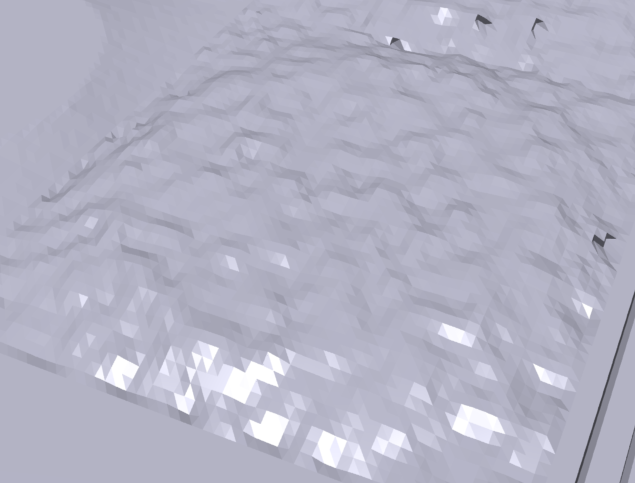}}%
\hfill
\subfloat[\textit{StairsUp}]{\label{4figs-c} \includegraphics[width=0.3\linewidth,height=2.5cm]{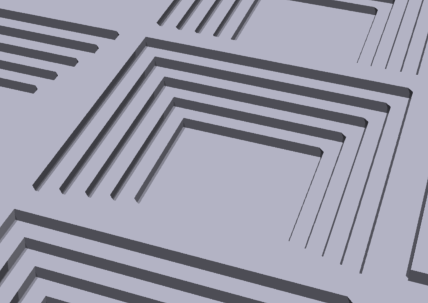}}%
\hfill
\subfloat[\textit{StairsDown}]{\label{4figs-d} \includegraphics[width=0.3\linewidth,height=2.5cm]{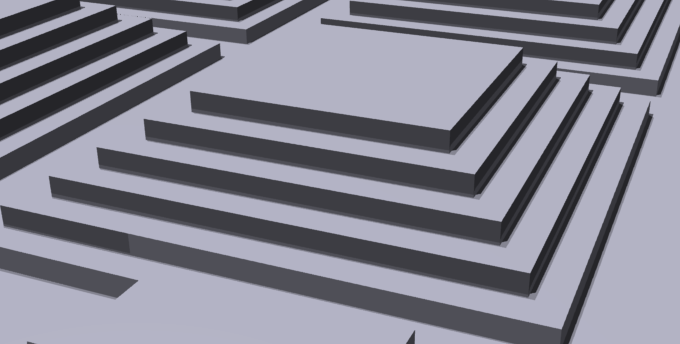}}%
\hfill
\subfloat[\textit{RandomObstacles}]{\label{4figs-e} \includegraphics[width=0.3\linewidth,height=2.5cm]{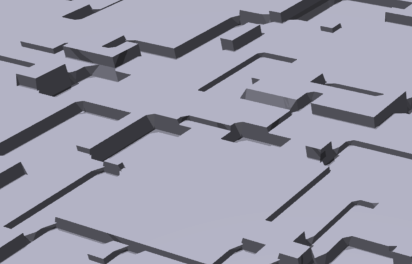}}%
\hfill
\subfloat[\textit{Wavy}]{\label{4figs-f} \includegraphics[width=0.3\linewidth,height=2.5cm]{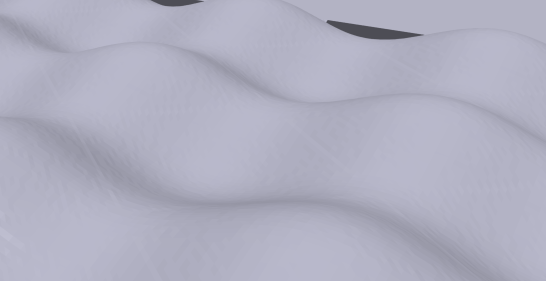}}%
\caption{Characteristic terrains considered in this work. Robot is trained only on terrains (a)-(e).}
\label{4figs}
\end{figure*}

Recently, there has been a surge of interest in learning-based control methods that address the problem of generalization of locomotion through non-flat unstructured terrains. Deep Reinforcement Learning (DRL) has been used to train robust control policies for unseen environments such as snow, mud and vegetation \cite{hwangbo2019learning, kumar2021rma, lee2020learning}. S. Chan et al. \cite{chen2022learning} resorted to learning  joint torques directly, while Shi et al. \cite{shi2022reinforcement} used a combined learning approach for the trajectory generator and joint target residues to successfully navigate through various terrains. Some works have used machine learning techniques for terrain perception \cite{kalakrishnan2011learning, magana2019fast, klamt2019towards}, while others have used DRL for terrain-aware locomotion \cite{heess2017emergence, peng2017deeploco,miki2022learning}. Wijmans et al. \cite{wijmans2019dd} leveraged a vast number of simulation samples to train a vision-based RL algorithm for unseen environments. Octavio V et al. \cite{magana2019fast} used Convolutional Neural Networks (CNNs) to learn foothold adaptation for unstructured terrains.

Despite the progress made in this area, there is still a gap in the literature regarding the evaluation of the effect of adding prismatic joints to a quadruped to traverse terrains with irregularities greater than foot reachability without resorting to dynamic gaits such as bounding and jumping.

\subsection{Contribution}
In contrast to the previous works discussed, which have focused on developing locomotion strategies without modifying the mechanical design of the quadruped, our work takes a novel approach by introducing a modification to the morphology. Specifically, we add an additional degree of freedom to each leg, which enables the quadruped not only to traverse through difficult terrains but also those that were previously infeasible for the nominal morphology to traverse without resorting to bounding gaits or jumping. This is because the feasible foothold would be out of the reachable region of the feet. Although approaches such as Sprowitz et al.\cite{sprowitz2013towards} have altered the mechanical design based on bio-inspired gaits, they do not add any extra degree of freedom, making our approach unique in comparison.
\vspace{10pt}

The remainder of this paper is organized as follows: In Section \ref{sys_des}, we briefly describe the modifications we made to a traditional quadrupedal robot model and provide a description of the types of terrain we use in simulation for training and evaluation. Section \ref{framework} outlines the reinforcement learning problem, describes the choice of state and action spaces, rewards, and training methodology. In Section \ref{Results}, we present the results of training and of our evaluation, measured by the Cost of Transport (CoT), of the trained policies on a variety of terrains for both robots with and without prismatic joints.

\section{System Design}\label{sys_des}
We have used Spot from Boston Dynamics as our quadruped base and modified the lower legs and foot links. A prismatic joint with a maximum range of $0.15$ m is added between the knee and foot of each leg. The robot has 16 actuated joints, 4 per leg:  3 revolute (hip adduction/abduction, hip flexion/extension and knee flexion/extension) and 1 linear (foot extension/retraction). The feet of the robot are simplified as spherical shaped links to simplify contact computations in simulation. The robot is shown in Fig.\ref{robot_image}. For training the quadrupeds for different terrains we have used environments inspired by\cite{Rudin2021LearningTW}. 
    The whole training environment consist five different terrains, difficulty level of the each terrain is increased from level 1 to 12 as described below:-
\begin{itemize}
    \item  \textit{SmoothSlope} [Fig.\ref{4figs-a}] :  With slope increasing from [$0.05$ - $0.4$] radians. 
    \item \textit{RoughSlope} [Fig.\ref{4figs-b}] : With slope increasing from [$0.05$ - $0.4$] radians with some added non uniformity.  
    \item \textit{StairsUp} [Fig.\ref{4figs-c}] : With stair height varying from [$0.05$ - $0.35$]m, and stair width varying as [$0.05$ - $0.3$]m.
    \item \textit{StairsDown} [Fig.\ref{4figs-d}] : With stair height varying from [$0.05$ - $0.35$]m, and stair width varying as [$0.05$ - $0.3$]m.
    \item \textit{RandomObstacles} [Fig.\ref{4figs-e}] : With obstacle height randomly raging from [$0.05$ - $0.35$]m.
\end{itemize}
For evaluating the learned policy we have used terrains discussed in \ref{Results} with highest difficulty levels.

\section{Reinforcement Learning Framework}\label{framework}
This section describes the methodology adopted for training our quadrupedal robot in simulation with Deep Reinforcement Learning. First, a basic definition of the problem is given followed by details about the observation and action spaces, the learning algorithm and other specific parameters.

\subsection{Definitions}
In the reinforcement learning setting, the decision making process of an agent is modelled as a Markov Decision Process (MDP). An MDP is a tuple $(\mathcal{S},\mathcal{A},\mathcal{P},\mathcal{R})$ of the state set $\mathcal{S}$, the action set $\mathcal{A}$ i.e. the space of decisions, the transition function $\mathcal{P}$ which describes how the choice of an action $a_{t}$ at timestep $t$ affects the state at the next timestep ${s_{t+1}}$ and is often given by a distribution $\mathcal{P}(s_{t+1}|s_{t},a_{t})$, and a reward function $\mathcal{R}(s_t,a_t,s_{t+1})$ representing the merit of an action. \\

The objective of an RL agent is to maximize the total expected reward over a time horizon. In Deep RL this is often done by finding a policy $\pi_{\theta}:[\mathcal{S,A]}\xrightarrow{}[0,1]$ parameterized by a parameter vector $\theta$. The optimal parameter $\theta^{*}$ is found by an RL algorithm that maximises the expected reward of trajectories resulting from acting according to policy $\pi_\theta$ and is given by:
\begin{equation}
\theta^{*}=\argmax_\theta \sum_{t=1}^{T-1}\mathbb{E}_{\{(s,a)\}_t\sim p_{\pi_\theta}(s_t,a_t)}\mathcal{R}(s_t,a_t,s_{t+1})
\end{equation}

\subsection{Action and Observation Space}

\begin{itemize}[label=\arabic*)]
    \item[1)] \textit{Action Space:} As described previously, we modify a Boston Dynamics Spot robot by adding additional linear actuators between the knee and the foot for each leg. Each leg now has 4 actuated joints. We choose to directly learn targets for these 16 joints which would then be tracked by actuator level controllers. 
    \item[2)] \textit{Observation Space:} Our observation space comprises of external commands stating the desired base velocities in the \textit{xy}-direction and an angular velocity about the \textit{z} axis, positions and velocities of all the actuated joints, previous actions for all the actuated joints, the gravity vector expressed in the robot's base frame, linear and angular velocities of the base also expressed in the base frame. Since we are not considering blind locomotion in this work, we sample a grid of measured terrain height around the current position of the robot. In the work we use a 21$\times$11 grid with $10cm$ size. In total we have a 291-D observation space.
\end{itemize}

\subsection{Rewards}
We define reward functions to encourage the robots to maintain a certain desired base height and orientation (allowing only \textit{yaw} motions) and follow the commanded \textit{xy} linear and angular velocities. Undesirable motions of the robot base, i.e vertical velocities and angular velocities along the robot's \textit{x} and \textit{y} axes are penalized. Smoothness is encouraged by penalizing actuator torques and changes in actuator targets. Collisions of the robot's upper and lower legs with the terrain are penalized while collision of the robot base with the terrain is considered terminal and the environment is reset. Similar to \cite{Rudin2021LearningTW}, we also add a reward term to encourage longer times between successive contact of each foot with the terrain. 
\subsection{Learning Algorithm and Policy Network}
We use the Proximal Policy Optimization algorithm \cite{schulman2017proximal} which is considered to be the go to on-policy algorithm for continuous state and action spaces. The algorithm learns two networks, a policy network $\pi_\theta$ and a critic network $V_\phi$ parameterized by $\theta$ and $\phi$ respectively. For our work we assume that both $\pi_\theta$ and  $V_\phi$ neural networks are fully connected with identical sizes.

At every learning iteration, the current policy is rolled out for a maximum of $T$ timesteps for $N$ episodes. The experience collected during the rollout is used to update both the critic network $V_\phi$ and policy/actor network $\pi_\theta$. Unlike vanilla policy gradient algorithms, PPO ensures that large KL-Divergence between the updated and the old policy is penalized by solving an unconstrained optimization problem. Table \ref{tab:ppo} gives the training parameters, PPO hyper-parameters and neural network dimensions.

\subsection{Training Methodology}
We use NVIDIA's Isaac Gym \cite{makoviychuk2021isaac} with PhysX as the simulation engine for its parallel simulation capability. The terrain is set up as a grid of 8m$\times$8m tiles where each tile is one of five of characteristic types\cite{Rudin2021LearningTW}.The difficulty of each type of terrain increases along the rows. At each training iteration \textit{xy} linear velocities and \textit{z} angular velocities are sampled uniformly from $[v_x^{min},v_x^{max}]$,$[v_y^{min},v_y^{max}]$ and $[\omega_z^{min},\omega_z^{max}]$ respectively. To ensure training stability and prevent catastrophic forgetting we employ a terrain curriculum\cite{lee2020learning} wherein each robot is preassigned a terrain type and difficulty level and depending on its performance with the current policy it is either promoted to the next level of its assigned terrain type or demoted to a lower level. For training, we simulate $n=2048$ robots in parallel meaning that at every learning iteration the current policy is rolled out simultaneously for $N=n=2048$ episodes. 
\begin{table}[ht]
\centering
\caption{PPO Hyper-parameters and NN shape}
\begin{tabular}{|c|c|c}
\cline{1-2}
No. of robots ($n=N$)                  & 2048                      \\ \cline{1-2}
Batch size    ($N\times T$)                      & 2048$\times$24                   \\ \cline{1-2}
Mini batch size                     & 2048$\times$6                    \\ \cline{1-2}
Number of learning epochs           & 5                         \\ \cline{1-2}
Entropy Coefficient                 & 0.01                      \\ \cline{1-2}
Discount Factor                     & 0.99                      \\ \cline{1-2}
Advantage Estimator Discount Factor & 0.95                      \\ \cline{1-2}
Desired KL-Divergence               & 0.01                      \\ \cline{1-2}
Units per hidden layer              & {[}512,256,128{]}         \\ \cline{1-2}
Activation                          & Exponential Linear Unit   \\ \cline{1-2}
Learning rate                       & adaptive                  \\ \cline{1-2}
\end{tabular}
\label{tab:ppo}
\end{table}

To make the learned policy resilient to external disturbances the robot base is pushed externally up to $1m/s$ in a random direction every 15 seconds. Noise is also added to the measurements of joint and base positions and velocities. For sim2real purposes, the friction coefficient of the terrain is randomised between $0.3-1.25$ every episode and the robot base mass is perturbed by $\pm5$ kilograms.

\section{Results and Discussion} \label{Results}
\subsection{Training}
\begin{figure}[ht]
\centerline{\includegraphics[scale=0.55]{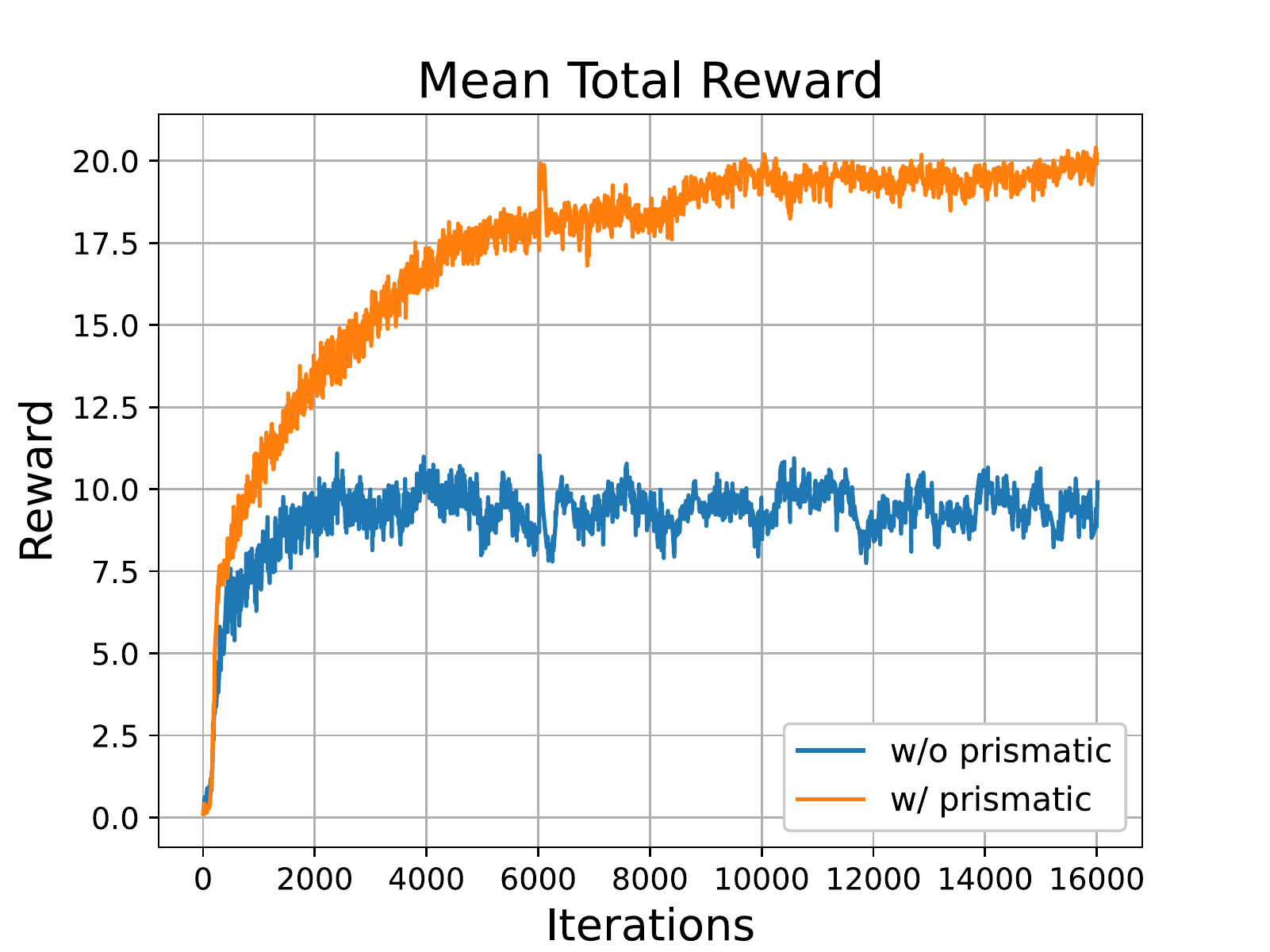}}
\caption{Mean total rewards during training}
\label{total_reward}
\end{figure}

The training was performed on a machine with an Intel Core-i9 processor, 32GB RAM and a 6GB Nvidia A3000 GPU. \\
For comparison, a policy was also trained for conventional quadruped morphology following an identical methodology (with a smaller observation space since there are fewer actuators) as presented earlier. The training time mean total reward for the two morphologies is shown in Fig. \ref{total_reward}. Note that the mean here is over the $2048$ robot simulation instances used to collect experience data for every learning iteration. 

\begin{figure}[ht]
\centerline{\includegraphics[scale=0.55]{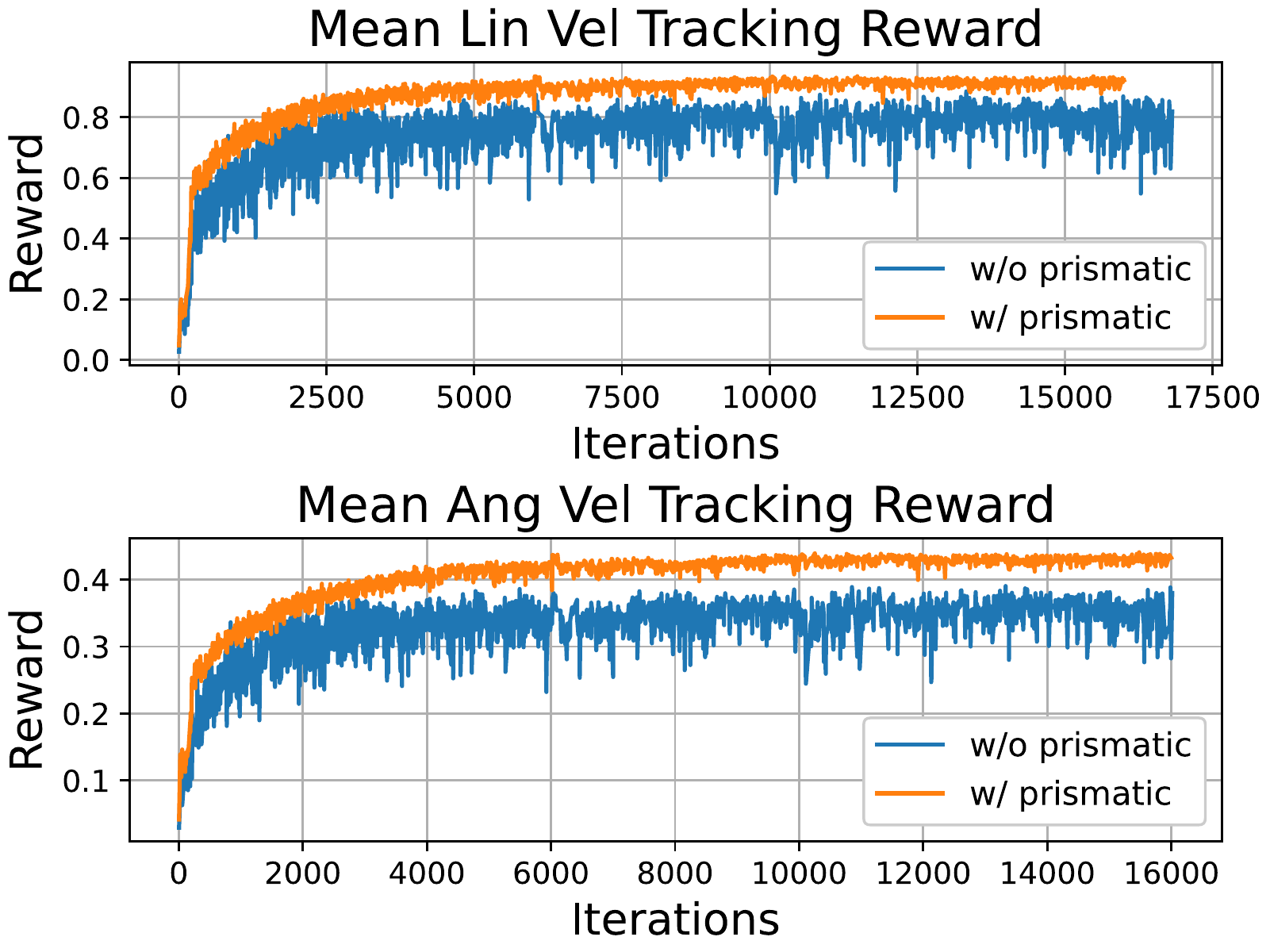}}
\caption{Mean velocity tracking rewards during training}
\label{vel_reward}
\end{figure}

The reward plot shows that with the training methodology adopted here, agents with additional prismatic joints are able to achieve significantly higher rewards. The rewards components for tracking velocity commands, as shown in Fig. \ref{vel_reward}, indicate that the learned policies for the agents with additional prismatic joints is on average better at tracking commanded base velocities. We use the exponential  function $f(v-v^*) = \exp(\frac {-\Vert v-v^*\Vert^2}{\sigma^2})$ as the velocity tracking reward, so perfect tracking corresponds to a reward of $1$.\\

\subsection{Evaluation}
We compare the two morphologies on different terrains on the basis of a dimensionless Cost of Transport metric as defined by

\begin{figure*}[tp]
\centering
\subfloat[]{\label{fig_comp_a} \includegraphics[width=0.23\linewidth, height=3cm]{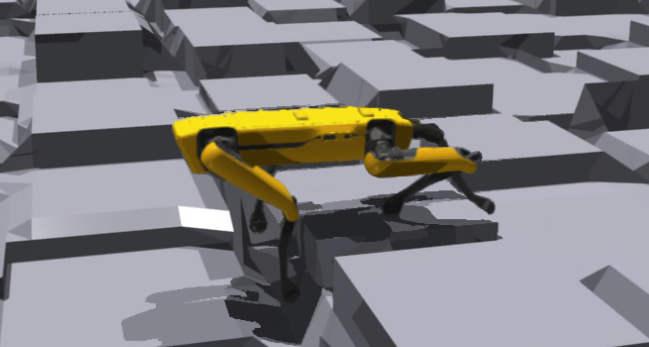}}
\subfloat[]{\label{fig_comp_b} \includegraphics[width=0.23\linewidth,height=3cm]{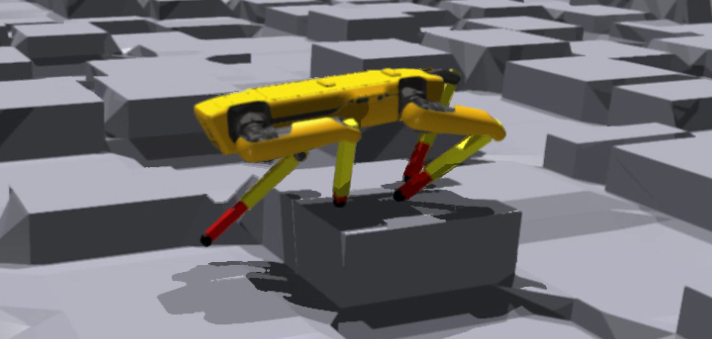}}
\subfloat[]{\label{fig_comp_c} \includegraphics[width=0.23\linewidth, height=3cm]{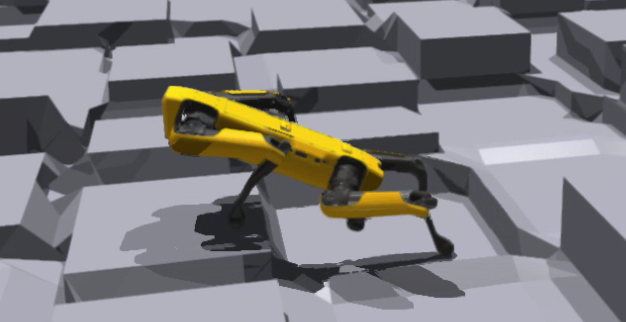}}
\subfloat[]{\label{fig_comp_d} \includegraphics[width=0.23\linewidth,height=3cm]{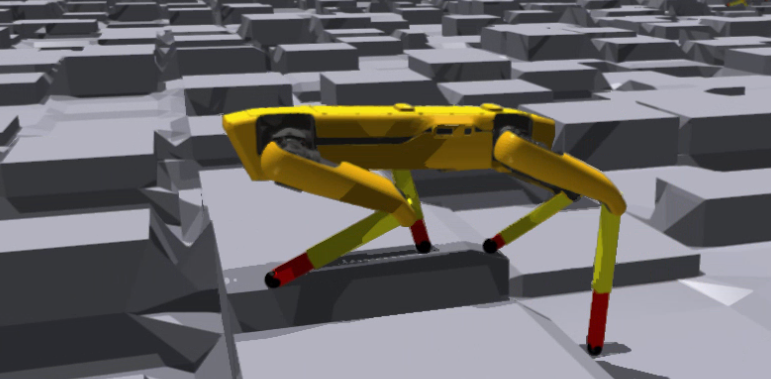}} \\
\subfloat[]{\label{fig_comp_e} \includegraphics[width=0.23\linewidth, height=3cm]{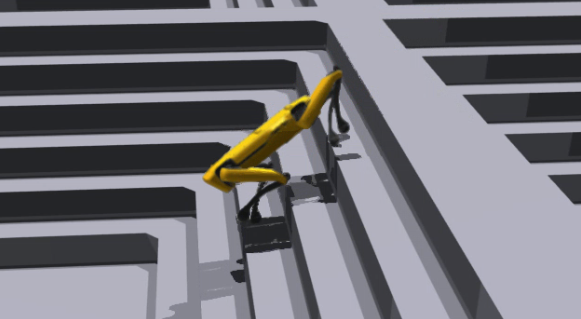}}
\subfloat[]{\label{fig_comp_f} \includegraphics[width=0.23\linewidth,height=3cm]{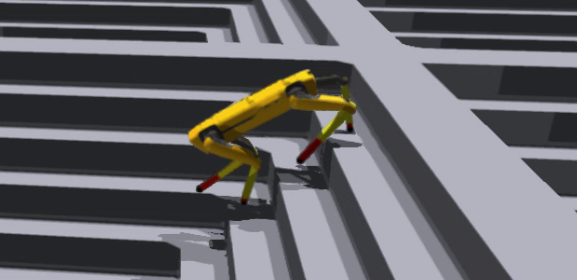}} 
\subfloat[]{\label{fig_comp_g} \includegraphics[width=0.23\linewidth, height=3cm]{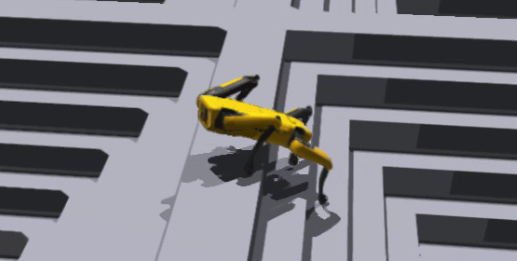}}
\subfloat[]{\label{fig_comp_h} \includegraphics[width=0.23\linewidth,height=3cm]{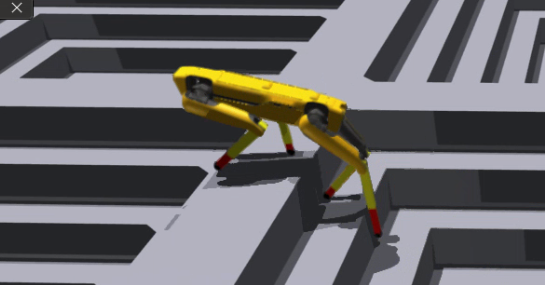}}
\caption{Quadruped taking advantage of prismatic links, while stepping up and down for different terrains.}

\label{comparision_figures} 
\end{figure*}

\begin{figure}[ht]
\centerline{\includegraphics[scale=0.24]{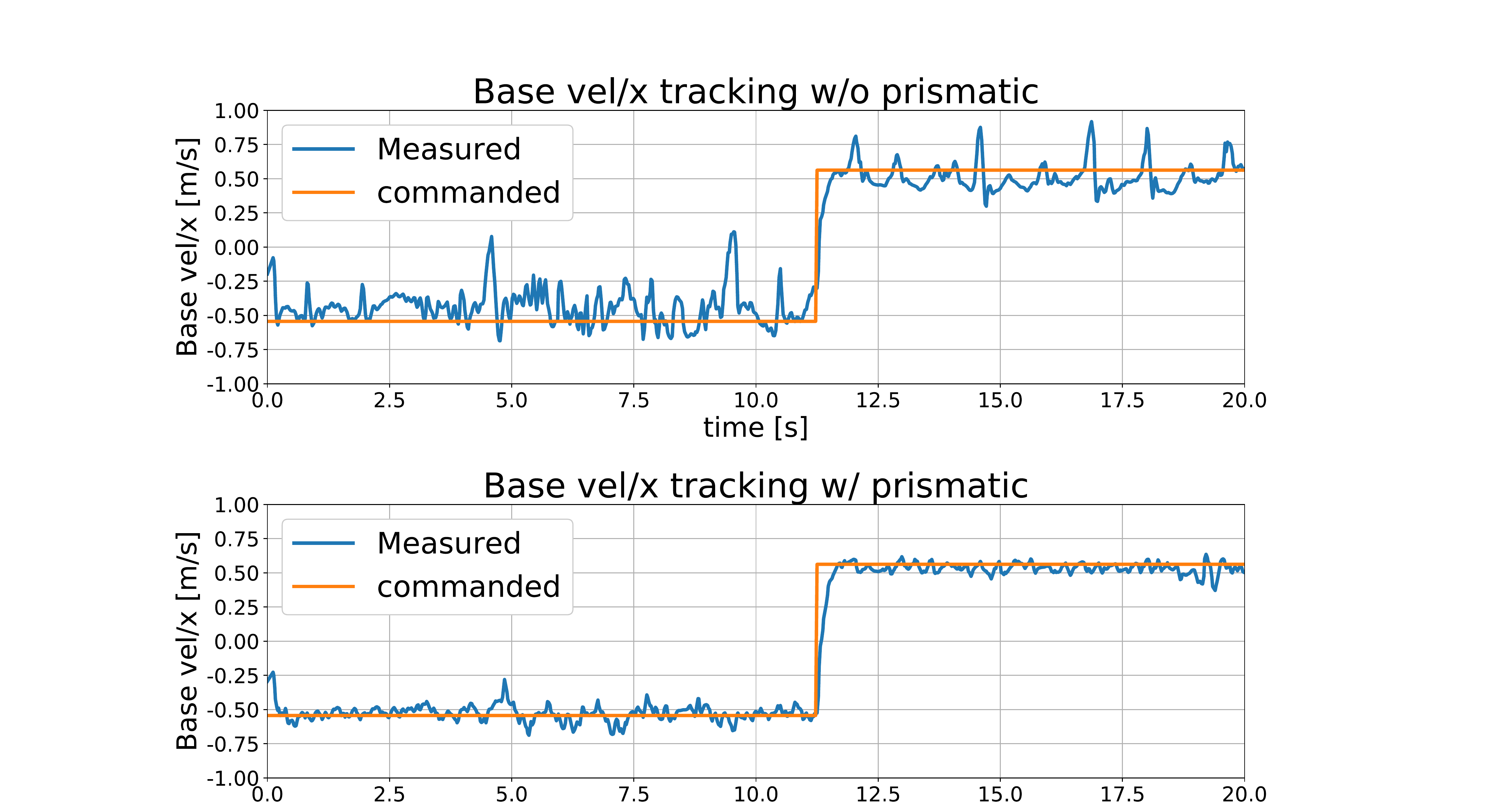}}
\caption{Measured and commanded base linear velocity in x}

\label{tracking_comparison_figure_1} 
\end{figure}

\begin{figure}[ht]
\centerline{\includegraphics[scale=0.24]{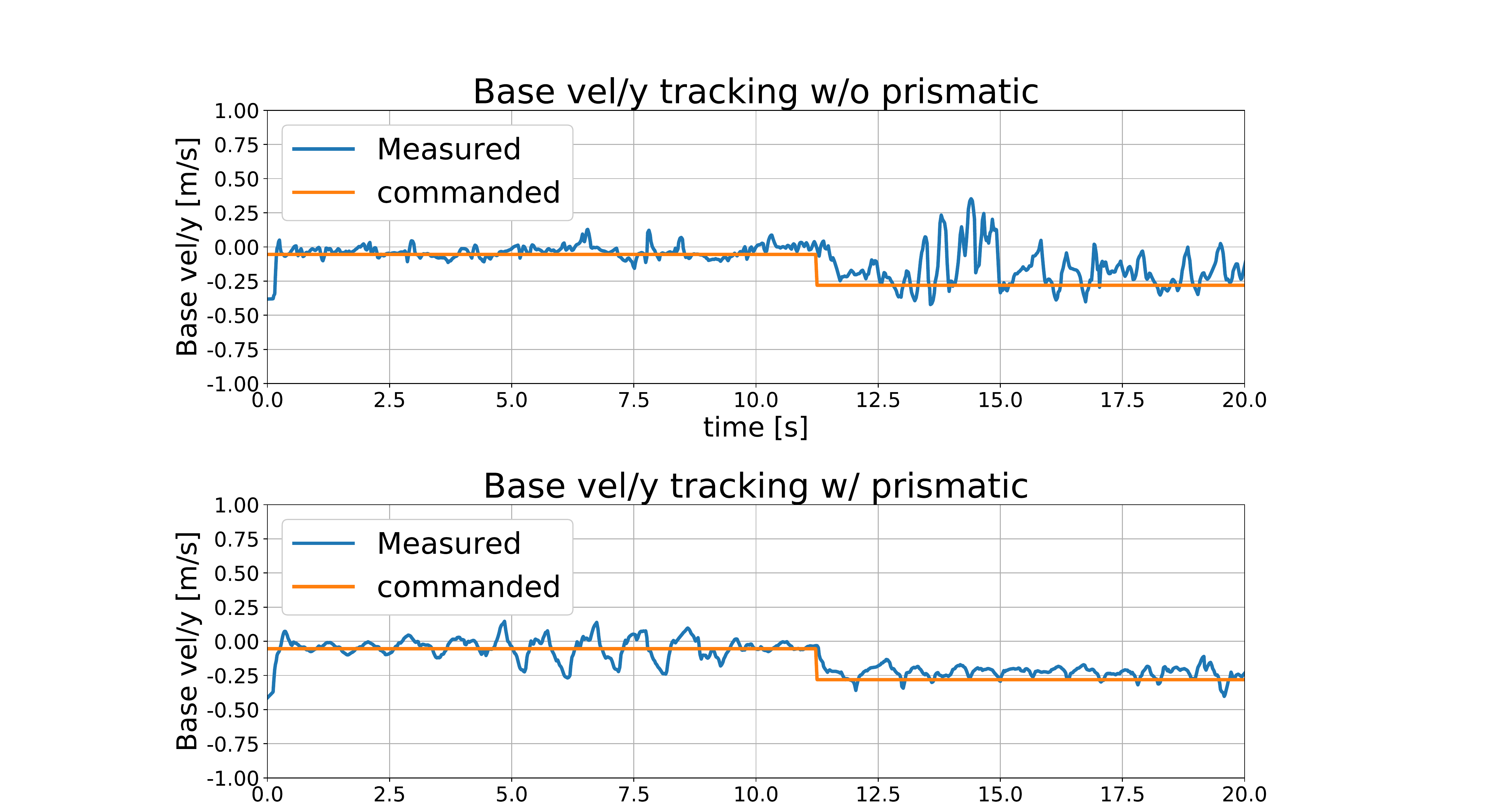}}
\caption{Measured and commanded base linear velocity in y}

\label{tracking_comparison_figure_2} 
\end{figure}

\begin{figure}[tbh]
\centerline{\includegraphics[scale=0.24]{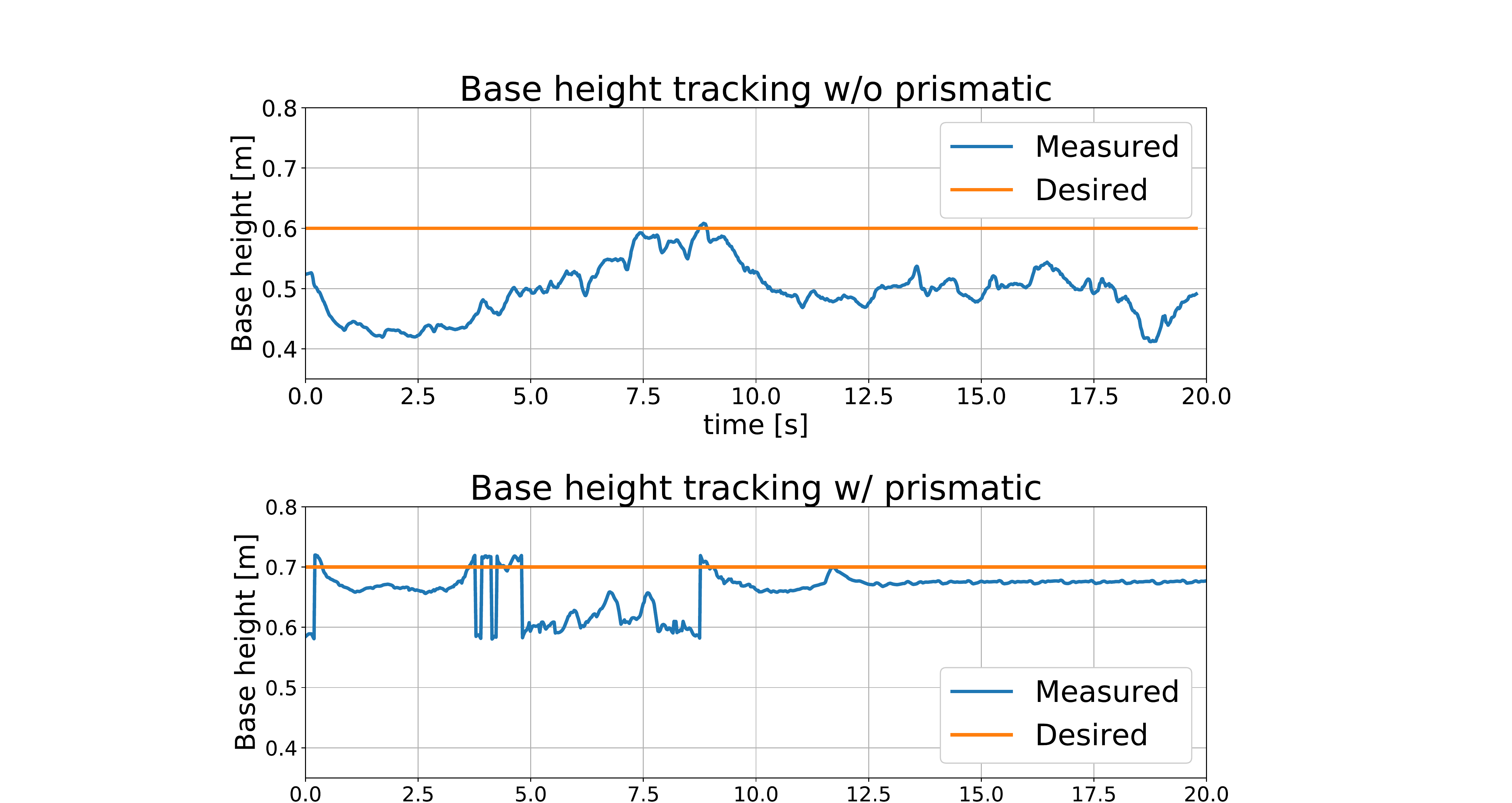}}
\caption{Measured and Desired base height}

\label{tracking_comparison_figure_3} 
\end{figure}

\begin{equation}
CoT=\sum_{actuators}\frac{[\tau\Dot{\theta}]^+_i}{MgV_b}
\label{eq:cot}
\end{equation}
where $[\tau\Dot{\theta}]^+_i$ is the positive mechanical power exerted by the $i$-th actuator, $M$ is the total mass, $g$ and $V_b$ are the gravitational acceleration and base locomotion speed respectively.

The terrains considered for the cost of transport study are the 5 types as mentioned in Sec. \ref{sys_des}, \textit{wavy} terrain as shown in Fig. \ref{4figs-f} and \textit{flat} terrain. Experiments are performed by spawning a robot in a 5$\times$5 tiling of a terrain type. The experiment is run for 20 seconds each. For the first 10 seconds robot is commanded a random pair of linear and angular velocities which remains constant within this duration. At 10 seconds another pair of velocities is given which remains constant till the end. The parameters of the terrains used in the experiments are:
\begin{itemize}[label=\arabic*)]
    \item[1)] \textit{StairsUp:} stair width: $0.3$m  , stair height: $0.35$m 
    \item[2)]  \textit{StairsDown:} stair width: $0.3$m  , stair height: $0.35$m
    \item[3)] \textit{SmoothSlope:} slope: $0.4$ radians
    \item[4)] \textit{RoughSlope:} slope: $0.4$ radians, uniform perturbation in height between [$0.05$ - $0.15$]m
    \item[5)]\textit{RandomObstacles:}  maximum obstacle height: $0.35$m , obstacle size between [$0.1$ - $0.8$]m
    \item[6)] \textit{Wavy:} no. of waves: $3$, amplitude: $0.35$m
\end{itemize}

These values correspond to the maximum terrain levels as mentioned in Sec. \ref{sys_des} except for the \textit{Wavy} terrain, parameters for which are chosen arbitrarily.

\begin{table}[b]
\centering
\caption{Average Cost of Transport from five 20sec runs}
\label{tab:cot_terrain}
\begin{tabular}{|c|cc|}
\hline
\multirow{2}{*}{\textit{\textbf{Terrain}}} & \multicolumn{2}{c|}{\textit{\textbf{Cost of Transport}}} \\ \cline{2-3} 
                                           & \multicolumn{1}{c|}{w/o prismatic}     & w/ prismatic    \\ \hline
\textit{StairsUp}                          & \multicolumn{1}{c|}{5.06}              & 1.84            \\ \hline
\textit{StairsDown}                        & \multicolumn{1}{c|}{3.33}              & 1.60             \\ \hline
\textit{SmoothSlope}                       & \multicolumn{1}{c|}{0.86}              & 1.04            \\ \hline
\textit{RoughSlope}                        & \multicolumn{1}{c|}{2.74}              & 1.90            \\ \hline
\textit{RandomObstacles}                   & \multicolumn{1}{c|}{2.88}              & 1.59            \\ \hline
\textit{Wavy}                              & \multicolumn{1}{c|}{1.17}                  & 1.40           \\ \hline
\textit{Flat}                              & \multicolumn{1}{c|}{0.52}                  & 0.72           \\ \hline
\end{tabular}
\end{table}

The detailed results are given in Table  \ref{tab:cot_terrain}. It is evident that use of prismatic joint significantly reduced the CoT in \textit{StairsUp}, \textit{StairsDown} and \textit{RandomObstacles} terrains. This could be explained with the help of Figs.  \ref{fig_comp_c}-\ref{fig_comp_h} which shows the actions taken by the learned policies of the two morphologies under similar circumstances. In the absence of the prismatic joints, the robot must use large motions in the hip joints to be able to find a suitable foot stance. This leads to higher mechanical power consumption due to the larger reflected inertia at the hip joints. On the other hand, with prismatic joints, the policy learns to use relatively smaller motions in the hip joints and the extension of the foot helps in finding feasible foot stance. The disparity is less evident in \textit{SmoothSlope} terrain and flips against the use of prismatic joints in the \textit{flat} terrain. This could be attributed to the contribution of the $4$ additional joints to the cost of transport in benign situations where those additional degrees of freedom might be useful. The added joints also assist the quadruped to keep track of the desired height (visible from Fig.\ref{tracking_comparison_figure_3}), while we evaluate the trained policy. At the same time the quadruped traverses the given terrains while maintaining the given linear velocity controls (visible from Fig.\ref{tracking_comparison_figure_1} and Fig.\ref{tracking_comparison_figure_2}).

\section{Conclusion and future work}
The results presented demonstrate the benefits of incorporating prismatic joints into the quadrupedal robot's design. Specifically, the robot with prismatic joints achieved higher rewards and demonstrated improved traversal of difficult terrains while maintaining commanded velocities and desired body height, resulting in a lower CoT compared to a more conventional design. However, a potential drawback of the DRL methodology used in this study is that the prismatic joints can be activated even in situations where they are not needed, leading to increased transportation costs. Future work could focus on developing more controlled actuation strategies to ensure the prismatic joints are only activated when necessary. Additionally, the transfer of the learned policy to perception-coupled hardware could be explored as an extension to this work.

\bibliographystyle{IEEEtran}
\bibliography{reference}

\begin{thebibliography}{10}
\providecommand{\url}[1]{#1}
\csname url@samestyle\endcsname
\providecommand{\newblock}{\relax}
\providecommand{\bibinfo}[2]{#2}
\providecommand{\BIBentrySTDinterwordspacing}{\spaceskip=0pt\relax}
\providecommand{\BIBentryALTinterwordstretchfactor}{4}
\providecommand{\BIBentryALTinterwordspacing}{\spaceskip=\fontdimen2\font plus
\BIBentryALTinterwordstretchfactor\fontdimen3\font minus
  \fontdimen4\font\relax}
\providecommand{\BIBforeignlanguage}[2]{{%
\expandafter\ifx\csname l@#1\endcsname\relax
\typeout{** WARNING: IEEEtran.bst: No hyphenation pattern has been}%
\typeout{** loaded for the language `#1'. Using the pattern for}%
\typeout{** the default language instead.}%
\else
\language=\csname l@#1\endcsname
\fi
#2}}
\providecommand{\BIBdecl}{\relax}
\BIBdecl

\bibitem{griffin2019footstep}
R.~J. Griffin, G.~Wiedebach, S.~McCrory, S.~Bertrand, I.~Lee, and J.~Pratt,
  ``Footstep planning for autonomous walking over rough terrain,'' in
  \emph{2019 IEEE-RAS 19th international conference on humanoid robots
  (humanoids)}.\hskip 1em plus 0.5em minus 0.4em\relax IEEE, 2019, pp. 9--16.

\bibitem{bledt2018contact}
G.~Bledt, P.~M. Wensing, S.~Ingersoll, and S.~Kim, ``Contact model fusion for
  event-based locomotion in unstructured terrains,'' in \emph{2018 IEEE
  International Conference on Robotics and Automation (ICRA)}.\hskip 1em plus
  0.5em minus 0.4em\relax IEEE, 2018, pp. 4399--4406.

\bibitem{focchi2018slip}
M.~Focchi, V.~Barasuol, M.~Frigerio, D.~G. Caldwell, and C.~Semini, ``Slip
  detection and recovery for quadruped robots,'' \emph{Robotics Research:
  Volume 2}, pp. 185--199, 2018.

\bibitem{camurri2017probabilistic}
M.~Camurri, M.~Fallon, S.~Bazeille, A.~Radulescu, V.~Barasuol, D.~G. Caldwell,
  and C.~Semini, ``Probabilistic contact estimation and impact detection for
  state estimation of quadruped robots,'' \emph{IEEE Robotics and Automation
  Letters}, vol.~2, no.~2, pp. 1023--1030, 2017.

\bibitem{kalakrishnan2010fast}
M.~Kalakrishnan, J.~Buchli, P.~Pastor, M.~Mistry, and S.~Schaal, ``Fast, robust
  quadruped locomotion over challenging terrain,'' in \emph{2010 IEEE
  International Conference on Robotics and Automation}.\hskip 1em plus 0.5em
  minus 0.4em\relax IEEE, 2010, pp. 2665--2670.

\bibitem{kolter2008control}
J.~Z. Kolter, M.~P. Rodgers, and A.~Y. Ng, ``A control architecture for
  quadruped locomotion over rough terrain,'' in \emph{2008 IEEE International
  Conference on Robotics and Automation}.\hskip 1em plus 0.5em minus
  0.4em\relax IEEE, 2008, pp. 811--818.

\bibitem{fukuoka2003adaptive}
Y.~Fukuoka, H.~Kimura, and A.~H. Cohen, ``Adaptive dynamic walking of a
  quadruped robot on irregular terrain based on biological concepts,''
  \emph{The International Journal of Robotics Research}, vol.~22, no. 3-4, pp.
  187--202, 2003.

\bibitem{righetti2008pattern}
L.~Righetti and A.~J. Ijspeert, ``Pattern generators with sensory feedback for
  the control of quadruped locomotion,'' in \emph{2008 IEEE International
  Conference on Robotics and Automation}.\hskip 1em plus 0.5em minus
  0.4em\relax IEEE, 2008, pp. 819--824.

\bibitem{hyun2014high}
D.~J. Hyun, S.~Seok, J.~Lee, and S.~Kim, ``High speed trot-running:
  Implementation of a hierarchical controller using proprioceptive impedance
  control on the mit cheetah,'' \emph{The International Journal of Robotics
  Research}, vol.~33, no.~11, pp. 1417--1445, 2014.

\bibitem{gay2013learning}
S.~Gay, J.~Santos-Victor, and A.~Ijspeert, ``Learning robot gait stability
  using neural networks as sensory feedback function for central pattern
  generators,'' in \emph{2013 IEEE/RSJ international conference on intelligent
  robots and systems}.\hskip 1em plus 0.5em minus 0.4em\relax Ieee, 2013, pp.
  194--201.

\bibitem{grandia2022perceptive}
R.~Grandia, F.~Jenelten, S.~Yang, F.~Farshidian, and M.~Hutter, ``Perceptive
  locomotion through nonlinear model predictive control,'' \emph{arXiv preprint
  arXiv:2208.08373}, 2022.

\bibitem{hwangbo2019learning}
J.~Hwangbo, J.~Lee, A.~Dosovitskiy, D.~Bellicoso, V.~Tsounis, V.~Koltun, and
  M.~Hutter, ``Learning agile and dynamic motor skills for legged robots,''
  \emph{Science Robotics}, vol.~4, no.~26, p. eaau5872, 2019.

\bibitem{kumar2021rma}
A.~Kumar, Z.~Fu, D.~Pathak, and J.~Malik, ``Rma: Rapid motor adaptation for
  legged robots,'' \emph{arXiv preprint arXiv:2107.04034}, 2021.

\bibitem{lee2020learning}
J.~Lee, J.~Hwangbo, L.~Wellhausen, V.~Koltun, and M.~Hutter, ``Learning
  quadrupedal locomotion over challenging terrain,'' \emph{Science robotics},
  vol.~5, no.~47, p. eabc5986, 2020.

\bibitem{chen2022learning}
S.~Chen, B.~Zhang, M.~W. Mueller, A.~Rai, and K.~Sreenath, ``Learning torque
  control for quadrupedal locomotion,'' \emph{arXiv preprint arXiv:2203.05194},
  2022.

\bibitem{shi2022reinforcement}
H.~Shi, B.~Zhou, H.~Zeng, F.~Wang, Y.~Dong, J.~Li, K.~Wang, H.~Tian, and
  M.~Q.-H. Meng, ``Reinforcement learning with evolutionary trajectory
  generator: A general approach for quadrupedal locomotion,'' \emph{IEEE
  Robotics and Automation Letters}, vol.~7, no.~2, pp. 3085--3092, 2022.

\bibitem{kalakrishnan2011learning}
M.~Kalakrishnan, J.~Buchli, P.~Pastor, M.~Mistry, and S.~Schaal, ``Learning,
  planning, and control for quadruped locomotion over challenging terrain,''
  \emph{The International Journal of Robotics Research}, vol.~30, no.~2, pp.
  236--258, 2011.

\bibitem{magana2019fast}
O.~A.~V. Magana, V.~Barasuol, M.~Camurri, L.~Franceschi, M.~Focchi, M.~Pontil,
  D.~G. Caldwell, and C.~Semini, ``Fast and continuous foothold adaptation for
  dynamic locomotion through cnns,'' \emph{IEEE Robotics and Automation
  Letters}, vol.~4, no.~2, pp. 2140--2147, 2019.

\bibitem{klamt2019towards}
T.~Klamt and S.~Behnke, ``Towards learning abstract representations for
  locomotion planning in high-dimensional state spaces,'' in \emph{2019
  International Conference on Robotics and Automation (ICRA)}.\hskip 1em plus
  0.5em minus 0.4em\relax IEEE, 2019, pp. 922--928.

\bibitem{heess2017emergence}
N.~Heess, D.~TB, S.~Sriram, J.~Lemmon, J.~Merel, G.~Wayne, Y.~Tassa, T.~Erez,
  Z.~Wang, S.~Eslami \emph{et~al.}, ``Emergence of locomotion behaviours in
  rich environments,'' \emph{arXiv preprint arXiv:1707.02286}, 2017.

\bibitem{peng2017deeploco}
X.~B. Peng, G.~Berseth, K.~Yin, and M.~Van De~Panne, ``Deeploco: Dynamic
  locomotion skills using hierarchical deep reinforcement learning,'' \emph{ACM
  Transactions on Graphics (TOG)}, vol.~36, no.~4, pp. 1--13, 2017.

\bibitem{miki2022learning}
T.~Miki, J.~Lee, J.~Hwangbo, L.~Wellhausen, V.~Koltun, and M.~Hutter,
  ``Learning robust perceptive locomotion for quadrupedal robots in the wild,''
  \emph{Science Robotics}, vol.~7, no.~62, p. eabk2822, 2022.

\bibitem{wijmans2019dd}
E.~Wijmans, A.~Kadian, A.~Morcos, S.~Lee, I.~Essa, D.~Parikh, M.~Savva, and
  D.~Batra, ``Dd-ppo: Learning near-perfect pointgoal navigators from 2.5
  billion frames,'' \emph{arXiv preprint arXiv:1911.00357}, 2019.

\bibitem{sprowitz2013towards}
A.~Spr{\"o}witz, A.~Tuleu, M.~Vespignani, M.~Ajallooeian, E.~Badri, and A.~J.
  Ijspeert, ``Towards dynamic trot gait locomotion: Design, control, and
  experiments with cheetah-cub, a compliant quadruped robot,'' \emph{The
  International Journal of Robotics Research}, vol.~32, no.~8, pp. 932--950,
  2013.

\bibitem{Rudin2021LearningTW}
N.~Rudin, D.~Hoeller, P.~Reist, and M.~Hutter, ``Learning to walk in minutes
  using massively parallel deep reinforcement learning,'' \emph{ArXiv}, vol.
  abs/2109.11978, 2021.

\bibitem{schulman2017proximal}
J.~Schulman, F.~Wolski, P.~Dhariwal, A.~Radford, and O.~Klimov, ``Proximal
  policy optimization algorithms,'' \emph{arXiv preprint arXiv:1707.06347},
  2017.

\bibitem{makoviychuk2021isaac}
V.~Makoviychuk, L.~Wawrzyniak, Y.~Guo, M.~Lu, K.~Storey, M.~Macklin,
  D.~Hoeller, N.~Rudin, A.~Allshire, A.~Handa \emph{et~al.}, ``Isaac gym: High
  performance gpu-based physics simulation for robot learning,'' \emph{arXiv
  preprint arXiv:2108.10470}, 2021.

\end{thebibliography}
\end{document}